\documentclass[lettersize,journal]{IEEEtran}
\usepackage{amsmath,amsfonts}
\usepackage{algorithmic}
\usepackage{algorithm}
\usepackage[normalem]{ulem}
\usepackage{array}
\usepackage[protrusion=true,expansion=true]{microtype}

\usepackage[caption=false,font=normalsize,labelfont=sf,textfont=sf]{subfig}
\usepackage{textcomp}
\usepackage{stfloats}
\usepackage{url}
\usepackage{verbatim}
\usepackage{graphicx}
\usepackage{cite}
\usepackage[colorlinks=true,linkcolor=black,citecolor=black,urlcolor=blue]{hyperref}

\begin{document}
\emergencystretch=4em

\title{\fontsize{21pt}{20pt}\selectfont Bidirectional Representational Alignment Between Biological and Artificial Neural Networks}

\author{Samuel Kostousov, Abhinn Kaushik, and Brokoslaw Laschowski,~\IEEEmembership{Member,~IEEE}
        
\thanks{This research was partially supported by the Schroeder Institute for Brain Innovation and Recovery.}
\thanks{S. Kostousov is with the Department of Physics, University of Toronto, Toronto, ON, Canada; (e-mail: samuel.kostousov@mail.utoronto.ca).}
\thanks{A. Kaushik is with the Department of Computer Science, University of Toronto, Toronto, ON, Canada; (e-mail: abhi.kaushik@mail.utoronto.ca).}
\thanks{B. Laschowski is with the Department of Mechanical and Industrial Engineering, University of Toronto, Toronto, ON, Canada; and the KITE Research Institute, University Health Network, Toronto, ON, Canada (e-mail: brokoslaw.laschowski@utoronto.ca).}}

\maketitle
\begin{abstract}
Recent work has shown that representational alignment between biological and artificial neural networks is asymmetric: model representations predict neural responses much better than neural responses predict model representations. This asymmetry raises the question of whether representational geometry contributes to bidirectional representational alignment. We hypothesized that steering representational geometry during training can systematically influence bidirectional alignment. To test this hypothesis, we developed a computational framework that integrates spectral regularization with bidirectional predictivity analyses. As an initial demonstration, we evaluated our framework using self-supervised contrastive vision models. Steering the spectral geometry of the learned representations substantially increased reverse predictivity with modest reductions in forward predictivity, yielding a 55\% relative improvement in bidirectional predictivity. These improvements were accompanied by reduced effective dimensionality and a reorganization of the shared representational subspace, within which forward and reverse predictivity became approximately symmetric at intermediate spectral exponents. Overall, these findings demonstrate that representational geometry can be systematically steered to modulate bidirectional representational alignment between biological and artificial neural networks.
\end{abstract}


\section{Introduction}
\IEEEPARstart{U}{nderstanding} how intelligence emerges from neural computations remains a central challenge in both machine learning and computational neuroscience. One promising approach is representational alignment, which compares the internal representations across different neural networks. By comparing the internal representations rather than external behavior alone, representational alignment provides a framework for studying the computational principles underlying intelligent systems and contributes to a deeper mechanistic understanding of intelligence \cite{bengio2013representation,KarDiCarlo2024ObjectRecognition,Schrimpf2020IntegrativeBenchmarking,gettingaligned,building_blocks,network_dissection,kornblith2019similarity}.

Artificial neural networks can develop representations that predict neural activity despite not being explicitly optimized using neural data \cite{og_predictivity,Schrimpf2020IntegrativeBenchmarking,ssl}. This observation is consistent with the Platonic Representation Hypothesis, which suggests that task optimization may recover fundamental principles of neural representation \cite{huh2024platonic,contprinc}. However, improvements in task performance do not always coincide with similar improvements in representational alignment, indicating that additional properties of learned representations may influence alignment \cite{imagenet_divergence,kornblith2019similarity}.

Representational alignment is most commonly evaluated using forward predictivity, which measures how accurately model representations predict neural activity in response to the same stimuli \cite{og_predictivity,neuroai_predictivity,gettingaligned}. Recently, \cite{reverse} introduced reverse predictivity, which measures how accurately neural activity predicts model representations. Using both metrics, they found a significant asymmetry: model representations predict neural responses much better than neural responses predict model representations. 

Previous research has shown that neural representations exhibit characteristic spectral geometry \cite{geom}, while regularizing representational geometry in artificial neural networks can alter the learned representations, model behavior, and alignment \cite{lossfunc,leno2026topological}. We hypothesized that steering representational geometry during learning can systematically influence bidirectional representational alignment.

\begin{figure*}[!h]
\centering
{\includegraphics[width=7.2 in]{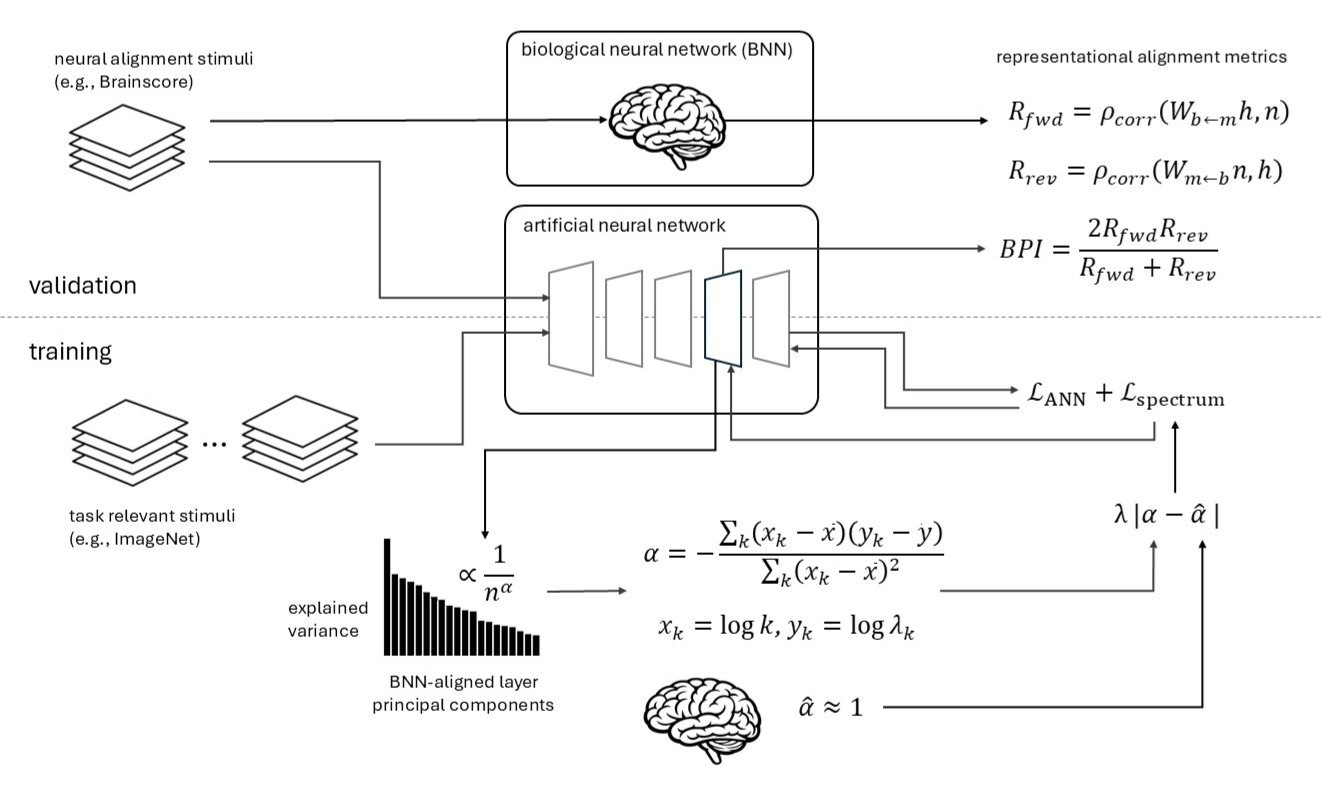}%
\hfil
\vspace{-1.5em}
\caption{Our computational framework for studying the relationship between representational geometry and bidirectional representational alignment between biological and artificial neural networks. During training, a standard learning objective is combined with spectral regularization to steer the spectral geometry of the learned representations by regularizing the spectral decay exponent. During validation, bidirectional predictivity is evaluated using forward and reverse predictivity on stimuli with neural data.}
\label{sys_diagram}}
\end{figure*}

To test this hypothesis, we developed a computational framework that combines controlled spectral steering during representation learning with bidirectional predictivity analyses (Fig.~\ref{sys_diagram}). As an initial demonstration, we applied our framework to self-supervised contrastive vision models and evaluated their alignment with representations in visual cortex. Our experiments show that steering representational geometry can substantially reduce forward--reverse predictivity asymmetry (i.e., mainly by increasing reverse predictivity) and that these changes are accompanied by reduced effective dimensionality and a reorganization of the shared representational subspace. Overall, this research provides a controlled framework for studying how properties of learned representational geometry influence bidirectional alignment between biological and artificial neural networks.

\section{Methods}
Our framework integrates spectral regularization for steering representational geometry with bidirectional predictivity analyses. As an initial demonstration, we implemented our framework using self-supervised contrastive vision models. Contrastive learning was selected because previous work \cite{reverse} has shown that forward--reverse predictivity asymmetry is particularly pronounced in this class of models, providing a suitable setting for evaluating changes in bidirectional representational alignment. Our source code is available at \url{https://github.com/skostousov/Bidirectional-Representational-Alignment} for reproducibility. 

\subsection{Dataset}
To evaluate our framework, we used ImageNet-1k for model training and a macaque inferotemporal (IT) cortex benchmark for evaluation. Specifically, we trained the model using a subset of ImageNet-1k \cite{imagenet} consisting of 100 images from each of the 1,000 classes, with balanced 80/20 training and validation splits. Class labels were not used during self-supervised training because the contrastive learning objective depends only on correspondences between independently augmented views of the same image. Bidirectional alignment was evaluated using an IT cortex benchmark \cite{Schrimpf2020IntegrativeBenchmarking,majaj2015}, comprising neural responses from 168 recording sites to 3,200 visual stimuli. Neural responses were averaged across 50 repeated presentations, and only sites with split-half reliability greater than 0.7 were retained. Neural data were used only for evaluation and did not influence model training, allowing forward and reverse predictivity to be interpreted as post hoc measures of bidirectional representational alignment.

\subsection{Model}
Using these datasets, we implemented our framework using SimCLR \cite{simclr}, which consists of a ResNet-50 encoder \cite{resnet} followed by a two-layer projection head that maps image representations into a 128-dimensional contrastive embedding space. We trained the baseline model using the standard SimCLR objective. In the regularized condition, the architecture, optimization procedure, and evaluation protocol were kept fixed, and only the training objective was modified to steer representational geometry through spectral regularization. 

Spectral regularization was applied to encoder.layer4.0.bn1, the first batch-normalization layer in the first bottleneck block of the encoder’s final residual stage (encoder.layer4.0). This bottleneck block has been reported to exhibit the strongest forward predictivity \cite{reverse}. Activations from encoder.layer4.0.bn1 were used for all subsequent analyses. For a batch of $B$ images, the activations were flattened into an activation matrix $A \in \mathbb{R}^{B \times D}$, where $D=C\times H\times W$ denotes the flattened feature dimension. 

\subsection{Training}
Using this model, we trained our framework following the standard SimCLR protocol \cite{simclr}. Two independently augmented views were generated from each image using the SimCLR augmentation pipeline. We trained the models for 200 epochs using stochastic gradient descent with momentum, cosine learning-rate decay with a 10-epoch warmup, and mixed-precision training when supported by hardware. All remaining hyperparameters followed the standard SimCLR implementation. We used the InfoNCE contrastive learning objective, which maximizes agreement between independently augmented views of the same image while minimizing agreement between views of different images \cite{hypersphereuniformity}. For a batch of $B$ images, the InfoNCE loss was computed as

\begin{equation}
    \mathcal{L}_{\text{InfoNCE}}
    =
    -\frac{1}{B}
    \sum_{j=1}^{B}
    \log
    \frac{
    \exp(\text{sim}(q_j,k_j^+)/\tau)
    }{
    \sum_{i=1}^{B}
    \exp(\text{sim}(q_j,k_i)/\tau)
    },
\end{equation}

where $\text{sim}(q_j,k_i)$ denotes cosine similarity between embeddings, $k_j^+$ is the positive pair corresponding to $q_j$, and $\tau$ is the temperature hyperparameter. In the spectral regularization experiments, we fixed $\tau = 0.2$ so that changes in representational geometry could be mainly attributed to spectral regularization. In addition to the contrastive loss, our framework computes a spectral regularization term from the activations of the selected encoder layer to steer representational geometry during learning. For each training batch, the activation matrix $A$ was used to estimate the eigenspectrum of the learned representations. The activation matrix was centered across samples, after which we performed a low-rank singular value decomposition using the first 40 principal components. The normalized explained variance spectrum was computed from singular values $S_i$ as

\begin{equation}
    \lambda_i =
    \frac{S_i^2}{\sum_j S_j^2}.
\end{equation}

Assuming that the explained variance spectrum follows an approximate power-law decay

\begin{equation}
    \lambda_i \propto i^{-\alpha},
\end{equation}

where the spectral exponent $\alpha$ was estimated by ordinary least squares on the log--log spectrum \cite{lossfunc}. The fit was computed over principal components 5--15, corresponding to the approximately linear portion of the spectrum. The spectral regularization term penalized deviations from a target spectral exponent $\hat{\alpha}$:

\begin{equation}
    \mathcal{L}_{\text{spectrum}}
    =
    \lambda |\alpha - \hat{\alpha}|,
\end{equation}

where $\lambda$ controls the strength of the spectral regularization. The total training objective was

\begin{equation}
    \mathcal{L}_{\text{total}}
    =
    \mathcal{L}_{\text{InfoNCE}}
    +
    \mathcal{L}_{\text{spectrum}}.
\end{equation}

Rather than directly optimizing bidirectional alignment, our framework steers the representational geometry by regularizing the spectral decay exponent. The target spectral exponent $\hat{\alpha}$ was fixed throughout each training run and initialized to approximately 1, consistent with eigenspectra reported for visual cortex \cite{geom}. Additional target values were also evaluated to characterize how representational geometry influences bidirectional alignment. To improve training stability, we introduced the spectral regularization after the first 10 training epochs.

\subsection{Inference}
Using the trained framework, we evaluated bidirectional representational alignment using forward and reverse predictivity. For both metrics, activations were extracted from the selected encoder layer for each stimulus with neural data. To compute forward explained variance, we fit ridge regression models from the model activations to each recorded neuron using five-fold cross-validation. Predictivity was quantified as the squared reliability-corrected Pearson correlation between predicted and observed responses on held-out data, where the raw correlation was divided by the geometric mean of the corresponding split-half reliabilities before squaring and converting to a percentage. To compute reverse explained variance, we fit ridge regression models from the neural responses to each model unit using the same cross-validation and reliability-correction procedure. Forward and reverse predictivity were computed by averaging the explained variance across neurons and model units, respectively. We also computed bidirectional predictivity, an asymmetry-penalizing composite metric introduced by \cite{reverse}:

\begin{equation}
BPI=
\frac{2R_{fwd}R_{rev}}
{R_{fwd}+R_{rev}},
\end{equation}

where $R_{fwd}$ and $R_{rev}$ denote the forward and reverse explained variance, respectively. To further characterize the mechanisms underlying bidirectional representational alignment, we identified the top 20\% of model units with the highest reverse predictivity, referred to as \emph{common units}, and recalculated forward and reverse predictivity using this subset. The same analysis was performed for the bottom 20\% of units, referred to as \emph{unique units}. We also evaluated the effective dimensionality of the learned representations. Using the normalized explained variance spectrum $\{\lambda_i\}$ computed from the stimuli used for evaluation, we calculated effective dimensionality as the participation ratio:

\begin{equation}
\mathrm{ED}
=
\frac{
\left(\sum_i \lambda_i\right)^2
}{
\sum_i \lambda_i^2
}.
\end{equation}

\subsection{Experiments}
Using these training and inference procedures, we evaluated our framework across a range of target spectral exponents $\hat{\alpha}$ and spectral regularization strengths $\lambda$. All remaining training hyperparameters were held fixed. Evaluations were performed after training due to high computational cost. All analyses were organized according to the measured spectral exponent $\alpha$ rather than the target exponent $\hat{\alpha}$ because lower regularization strengths did not always achieve the intended target value. This allowed us to study how steering representational geometry influenced bidirectional representational alignment independent of the nominal optimization target. The measured spectral exponent provided a more meaningful explanatory variable than the nominal regularization target. Models were grouped into measured spectral-exponent bins of $0.3$--$0.5$, $0.5$--$0.8$, $0.8$--$1.2$, and $1.2$--$1.6$ for analysis, with the unregularized baseline included separately. Combinations of spectral-exponent bins and regularization strengths containing fewer than three models were excluded.

\section{Results}

\begin{figure}[!t]
    \centering
    \includegraphics[width=\linewidth]{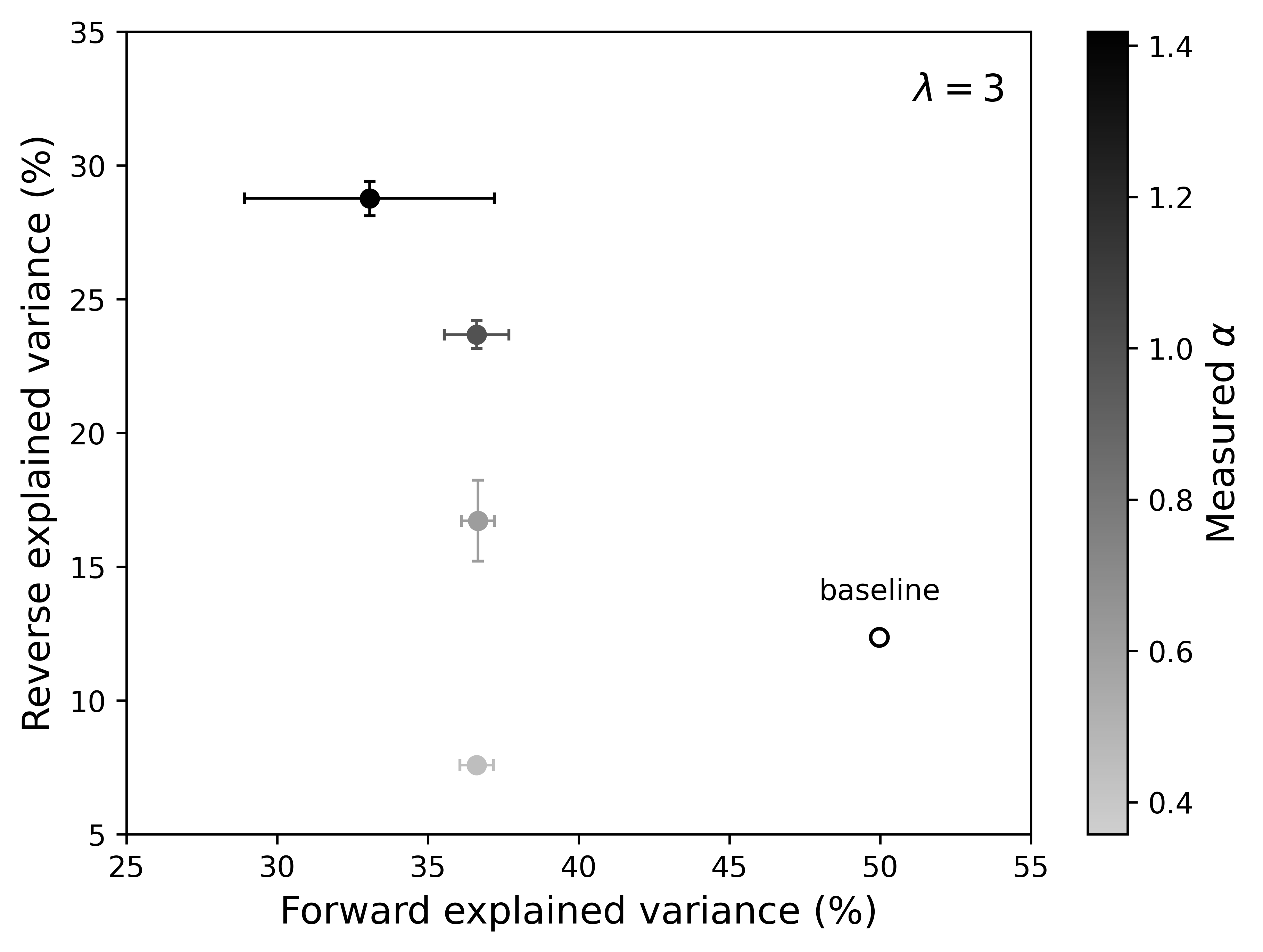}
    \caption{Forward and reverse explained variance across the measured spectral exponents for $\lambda=3$. The white point is the unregularized baseline. Error bars represent the median absolute deviation.}
    \label{fig:rf3}
\end{figure}

\begin{figure}[!t]
    \centering
    \includegraphics[width=\linewidth]{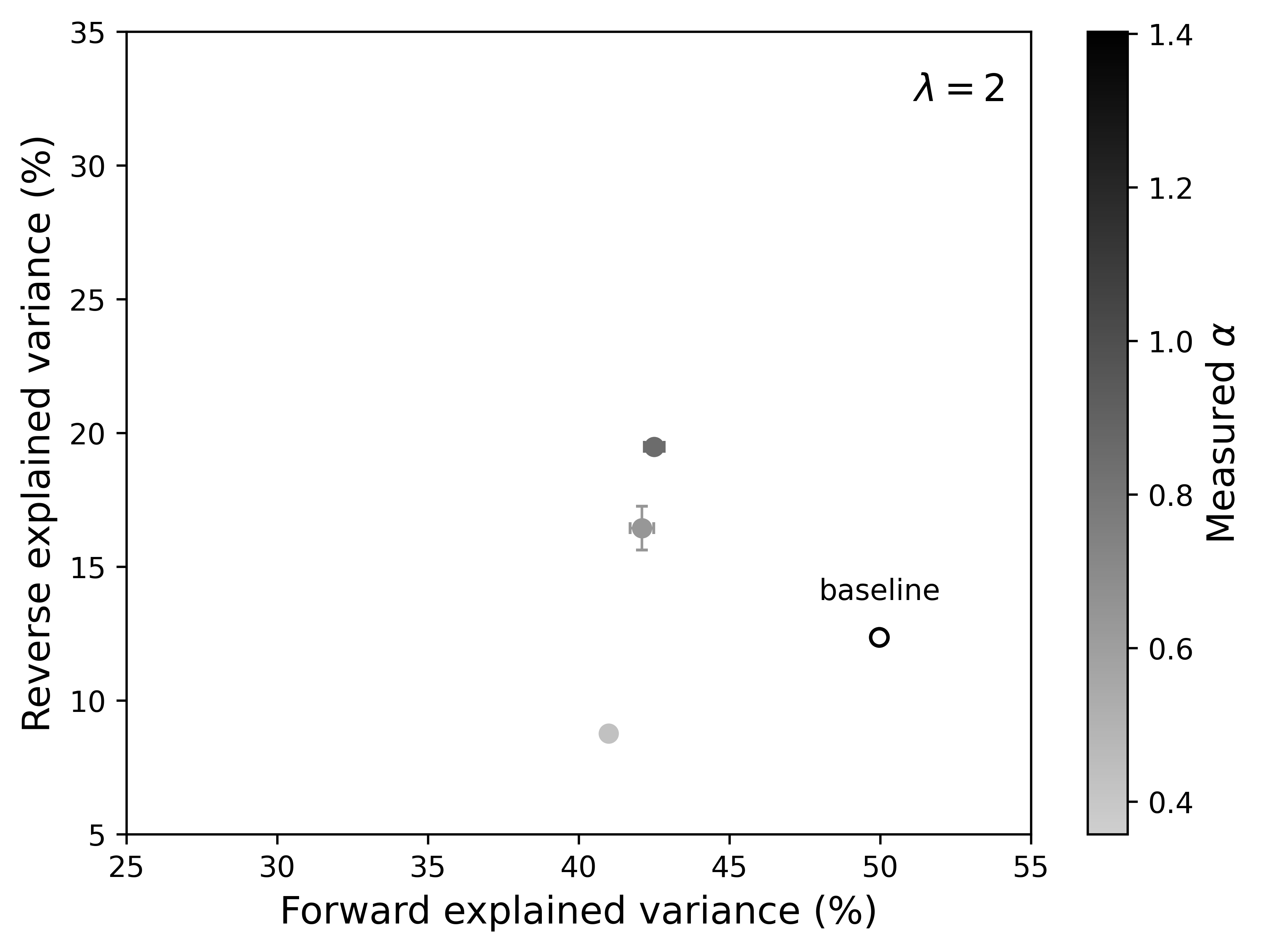}
    \caption{Forward and reverse explained variance across the measured spectral exponents for $\lambda=2$. The white point is the unregularized baseline. Error bars represent the median absolute deviation.}
    \label{fig:rf2}
\end{figure}

First, we confirmed that spectral regularization was able to steer the representational geometry. The measured spectral exponents increased with the target spectral exponent, and stronger regularization produced closer agreement between the measured and target values. We observed no systematic differences in agreement between the measured and target values across the evaluated targets. Only target values larger than the baseline were evaluated. These results demonstrate that spectral regularization can systematically steer the spectral geometry of the learned representations.

\begin{figure}[!t]
    \centering
    \includegraphics[width=\linewidth]{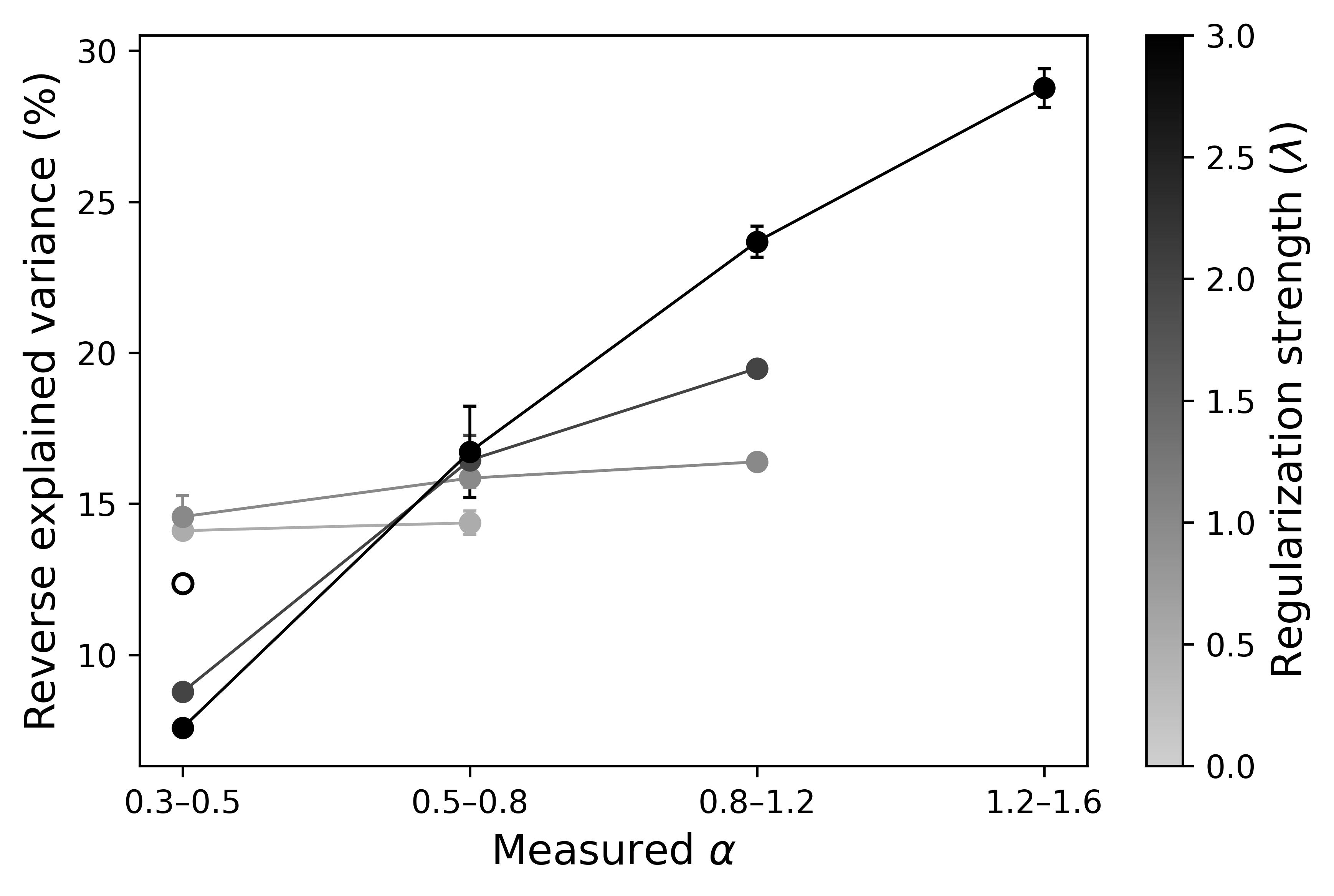}
    \caption{Reverse explained variance as a function of the measured spectral exponent $\alpha$ and regularization strength $\lambda$. The white point is the unregularized baseline. Error bars represent the median absolute deviation.}
    \label{fig:rlambda}
\end{figure}

\begin{figure}[!t]
    \centering
    \includegraphics[width=\linewidth]{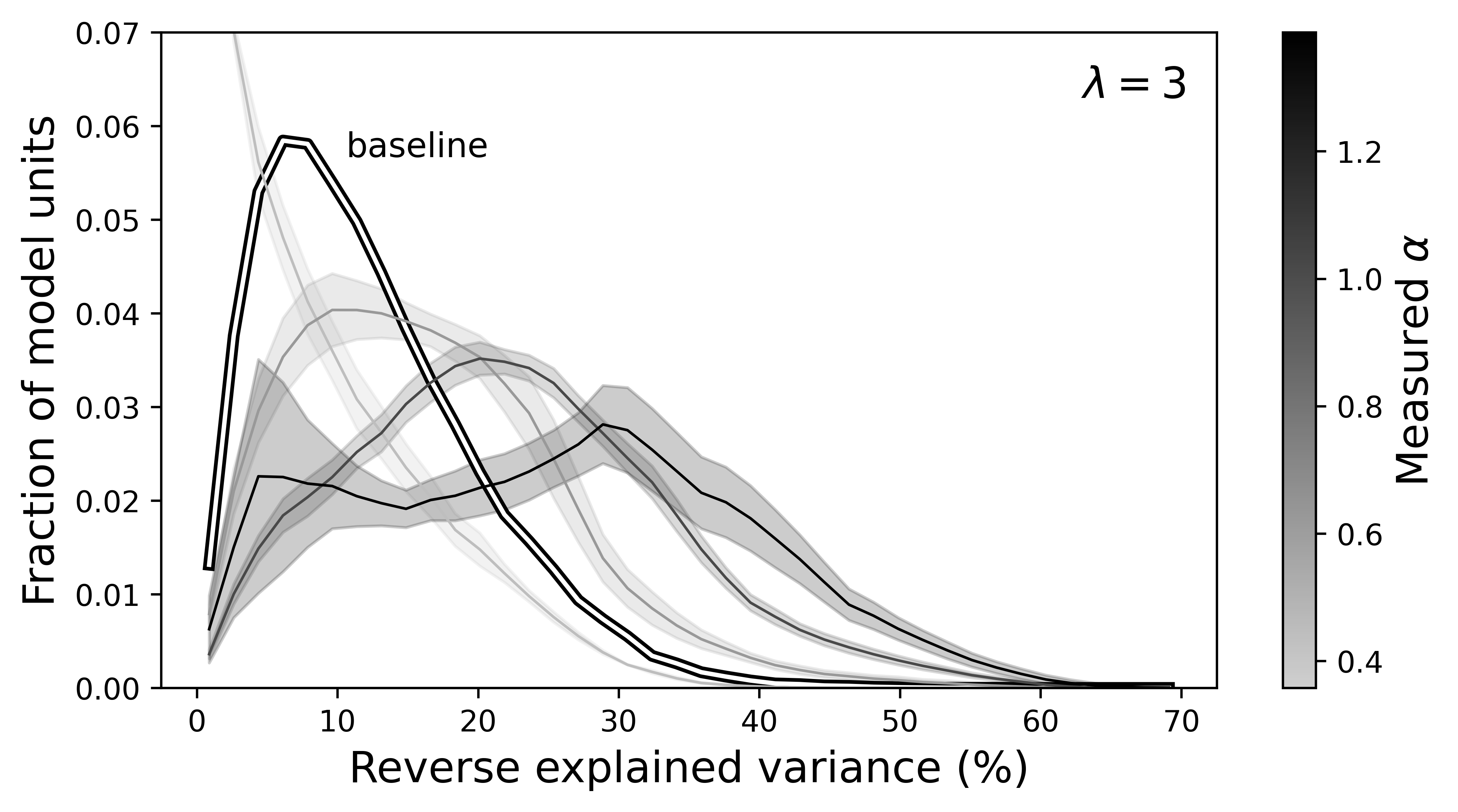}
    \caption{Distributions of reverse explained variance across model units for $\lambda=3$. The white distribution is the unregularized baseline. Error bands are the standard error of the mean.}
    \label{fig:rhist3}
\end{figure}

We then evaluated how steering representational geometry influenced reverse, forward, and bidirectional predictivity. Reverse predictivity increased with the measured spectral exponent at $\lambda=3$ (Fig.~\ref{fig:rf3}). Relative to the unregularized baseline, models with larger measured $\alpha$ exhibited higher reverse explained variance, with the largest improvements observed in the highest-$\alpha$ bin, where reverse explained variance increased from $12\%$ to $29\%$. A similar trend was observed at $\lambda=2$ (Fig.~\ref{fig:rf2}). An exception was the worse-than-baseline reverse predictivity for models regularized to a similar spectral exponent as the unregularized baseline. As shown in Fig.~\ref{fig:rlambda}, reverse predictivity increased with the measured spectral exponent and regularization strength. Larger measured $\alpha$ were generally associated with larger reverse explained variance across the population of model units, indicating that improvements were not confined to a small subset (Fig.~\ref{fig:rhist3}).

\begin{figure}[!t]
    \centering
    \includegraphics[width=\linewidth]{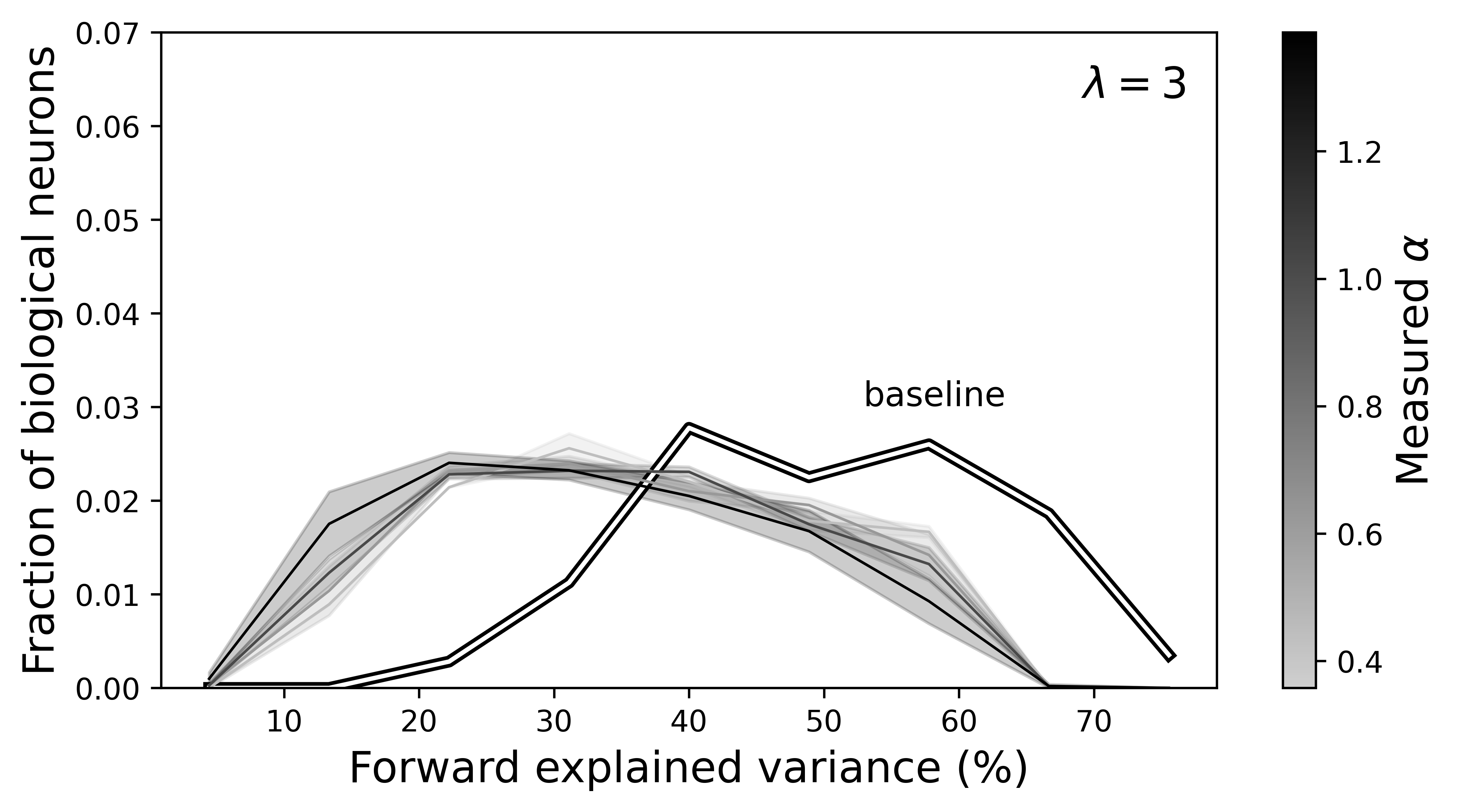}
    \caption{Distributions of forward explained variance across recorded neurons for models with $\lambda=3$. The white distribution represents the unregularized baseline. Error bands are the standard error of the mean.}
    \label{fig:fhist3}
\end{figure}

\begin{figure}[!t]
    \centering
    \includegraphics[width=\linewidth]{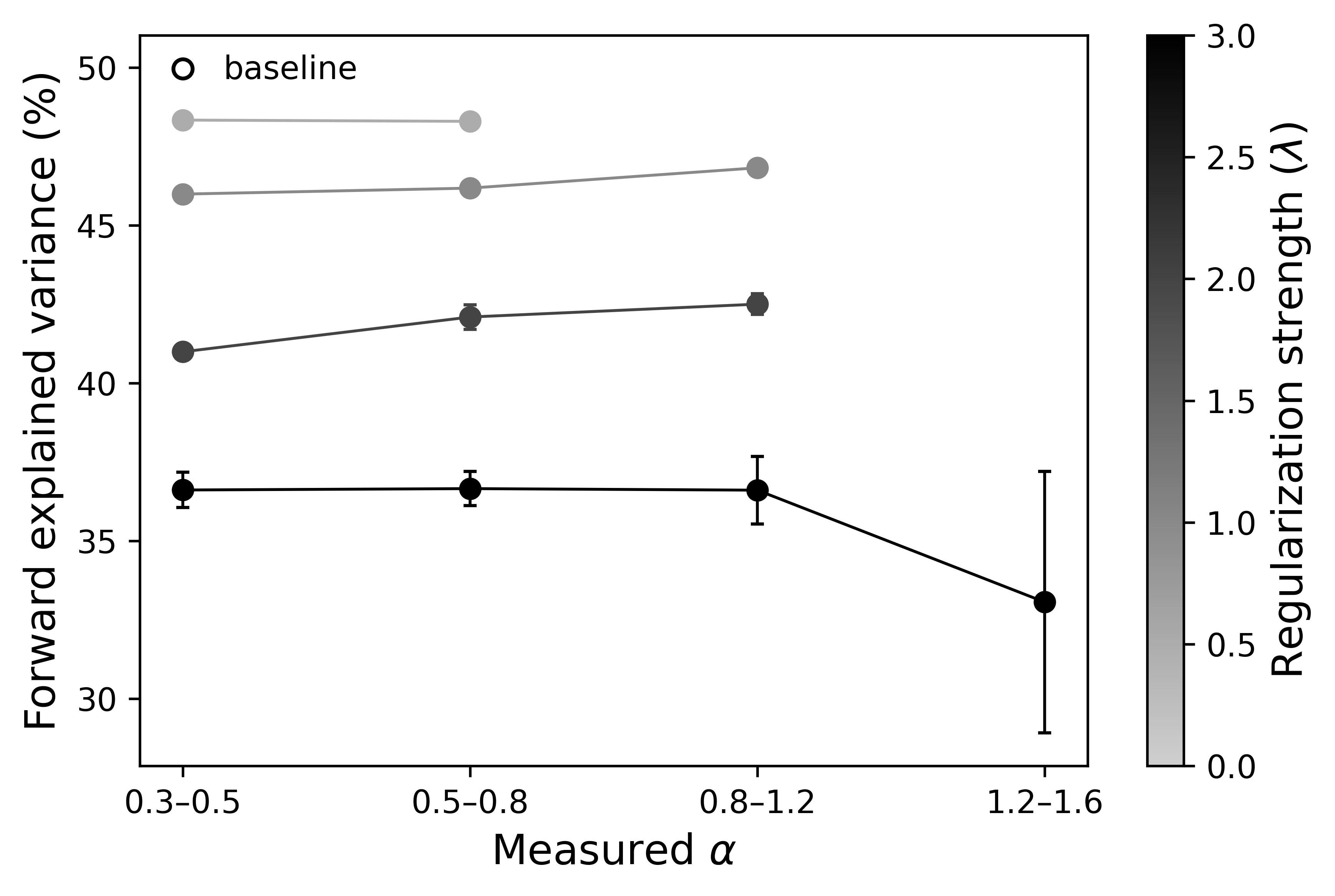}
    \caption{Forward explained variance as a function of the measured spectral exponent $\alpha$ and spectral regularization strength $\lambda$. The white point is the unregularized baseline. Error bars are the median absolute deviation.}
    \label{fig:flambda}
\end{figure}

Forward predictivity exhibited a different relationship. Relative to the baseline, forward explained variance generally decreased under spectral regularization (Figs.~\ref{fig:rf3} and \ref{fig:rf2}). Unlike reverse predictivity, forward predictivity varied comparatively less across the measured $\alpha$ at a fixed regularization strength. The distributions of forward explained variance were consistent with these findings (Fig.~\ref{fig:fhist3}). Regularization shifted the distributions toward lower forward explained variance, with relatively small differences across the measured spectral exponents. Grouping models by measured $\alpha$ suggested that forward predictivity was more strongly associated with spectral regularization strength than with the measured spectral exponent (Fig.~\ref{fig:flambda}). These findings suggest that larger spectral exponents are primarily associated with increased reverse predictivity, while stronger regularization enables higher $\alpha$ but is also associated with reduced forward predictivity.

\begin{figure}
    \centering
    \includegraphics[width=\linewidth]{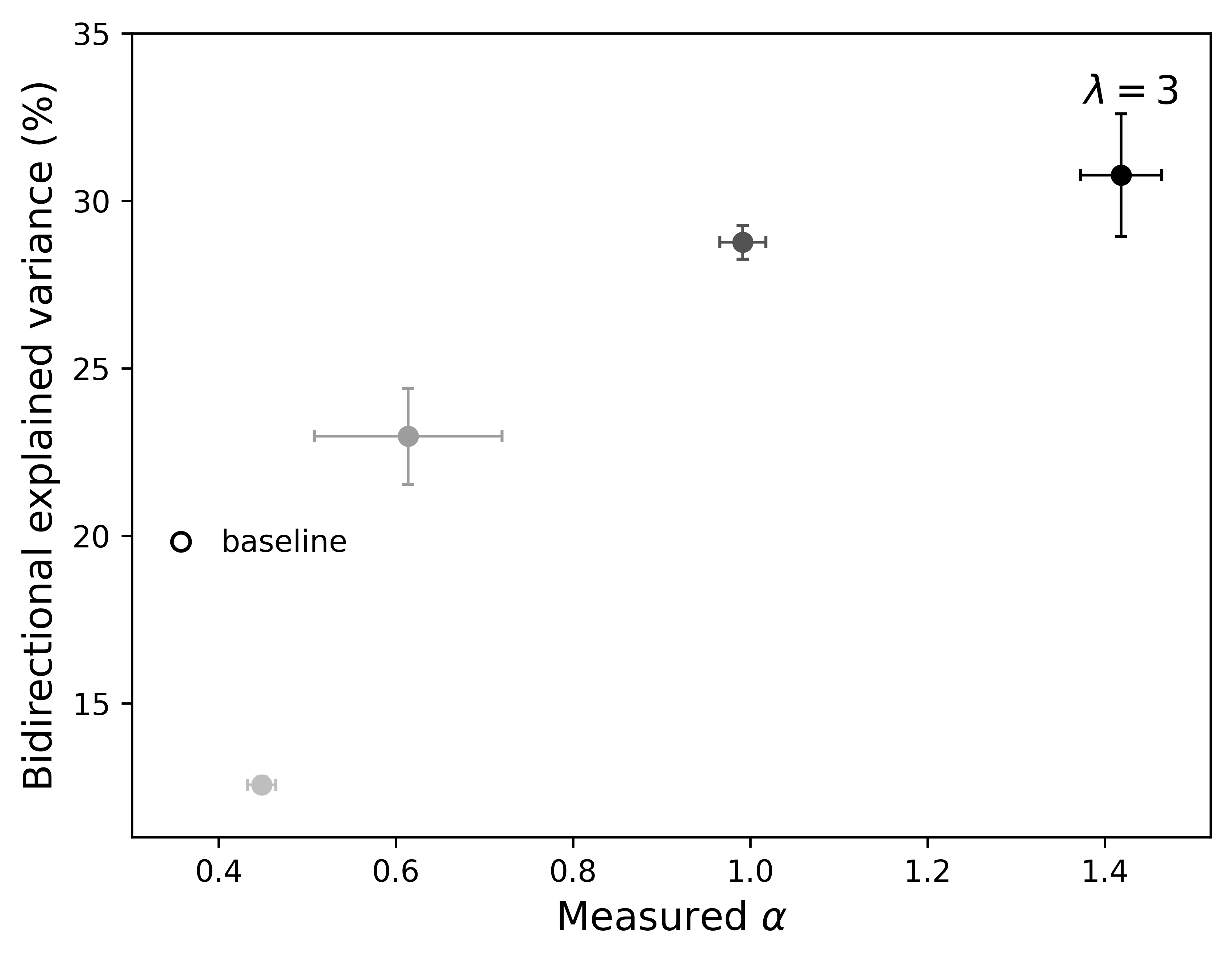}
    \caption{Bidirectional predictivity as a function of the measured spectral exponent $\alpha$ for models with $\lambda=3$. The white point is the unregularized baseline. Error bars represent the median absolute deviation.}
    \label{fig:bpi}
\end{figure}

To evaluate the net effect of this trade-off, we computed bidirectional predictivity, which jointly quantifies forward and reverse predictivity while penalizing asymmetry between them. Bidirectional predictivity increased with the measured spectral exponent (Fig.~\ref{fig:bpi}). Although steering the representational geometry reduced forward predictivity, it substantially increased reverse predictivity, reducing asymmetry between the two directions and increasing bidirectional predictivity from approximately 20\% to 31\%, corresponding to a relative improvement of 55\%. Together, these findings demonstrate that spectral regularization can systematically influence bidirectional representational alignment. The accompanying changes in representational geometry are consistent with our hypothesis that representational geometry contributes to this improvement.

\begin{figure}[!t]
    \centering
    \includegraphics[width=\linewidth]{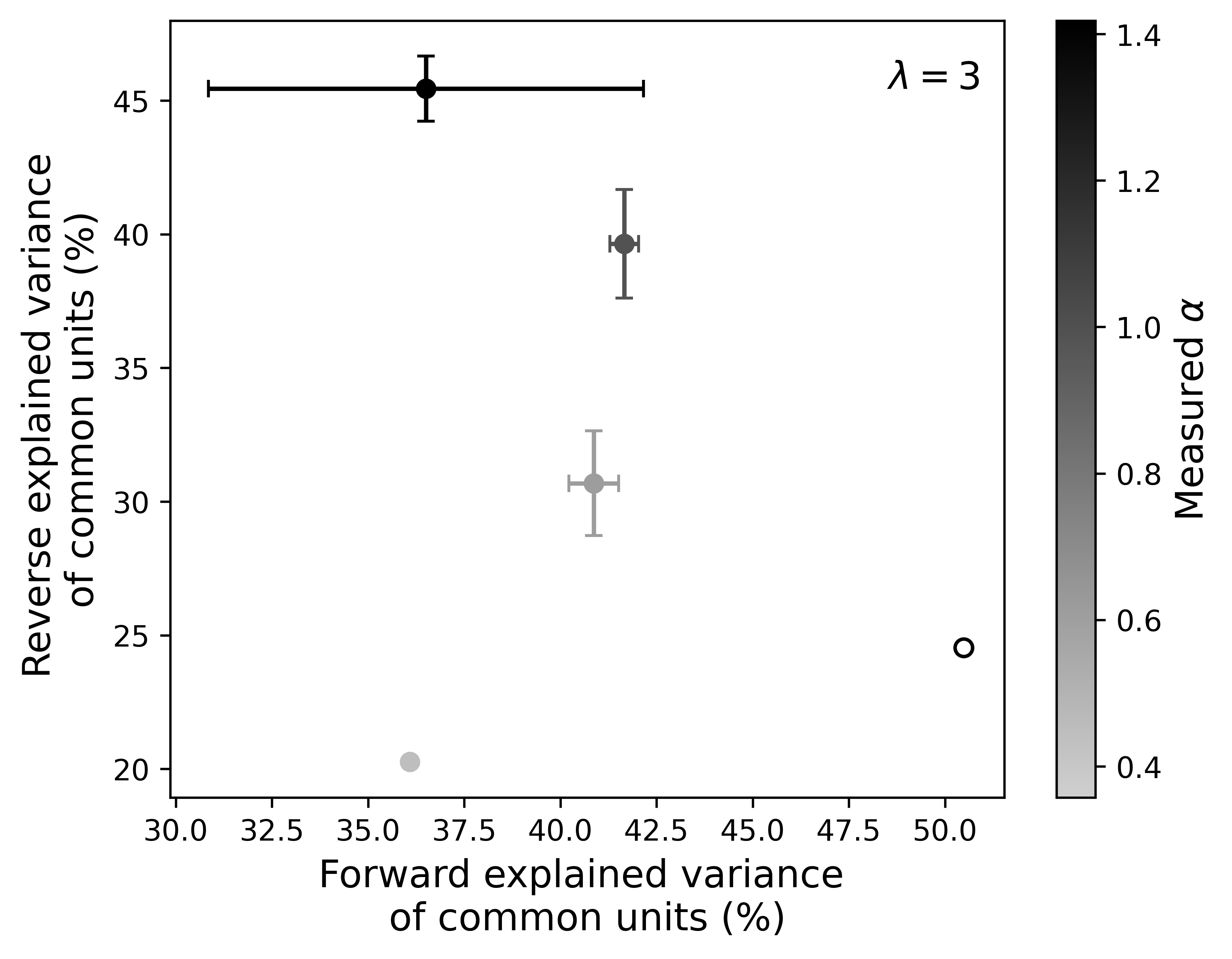}
    \caption{Forward and reverse explained variance for common model units (top 20\% ranked by reverse predictivity). The white point denotes the unregularized baseline. Error bars are the median absolute deviation.}
    \label{fig:common_units}
\end{figure}

\begin{figure}[!t]
    \centering
    \includegraphics[width=\linewidth]{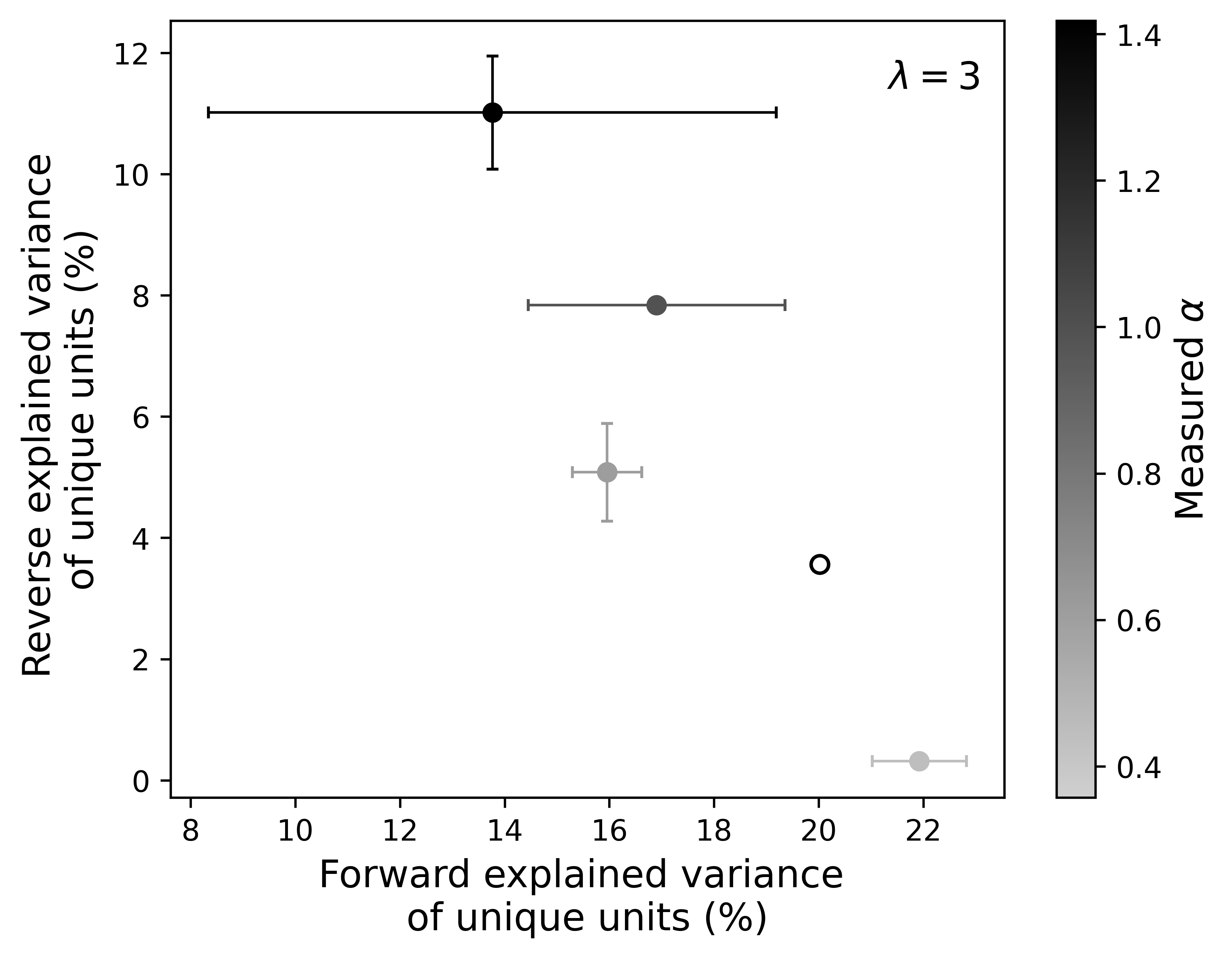}
    \caption{Forward and reverse explained variance for unique model units (bottom 20\% ranked by reverse predictivity). The white point denotes the unregularized baseline. Error bars are the median absolute deviation.}
    \label{fig:unique_units}
\end{figure}

We further studied bidirectional predictivity by separately analyzing common and unique model units. Here, common units are the top 20\% of model units ranked by reverse predictivity, whereas unique units are the bottom 20\%. Because common units are defined according to reverse predictivity, analyzing forward predictivity within this subset allowed us to characterize how representations most strongly shared with neural activity changed under spectral regularization. As shown in Fig.~\ref{fig:common_units}, reverse predictivity among common units increased with the measured spectral exponent, reaching $45\%$ in the highest $\alpha$ bin while forward predictivity fell to $37\%$. Unique units exhibited a similar trend but with substantially smaller gains in reverse predictivity (Fig.~\ref{fig:unique_units}). These results indicate that spectral regularization preferentially reorganizes the representational structure most strongly shared between artificial and biological neural networks, with the largest improvements in reverse predictivity occurring among the common units.

\begin{figure}[!t]
    \centering
    \includegraphics[width=\linewidth]{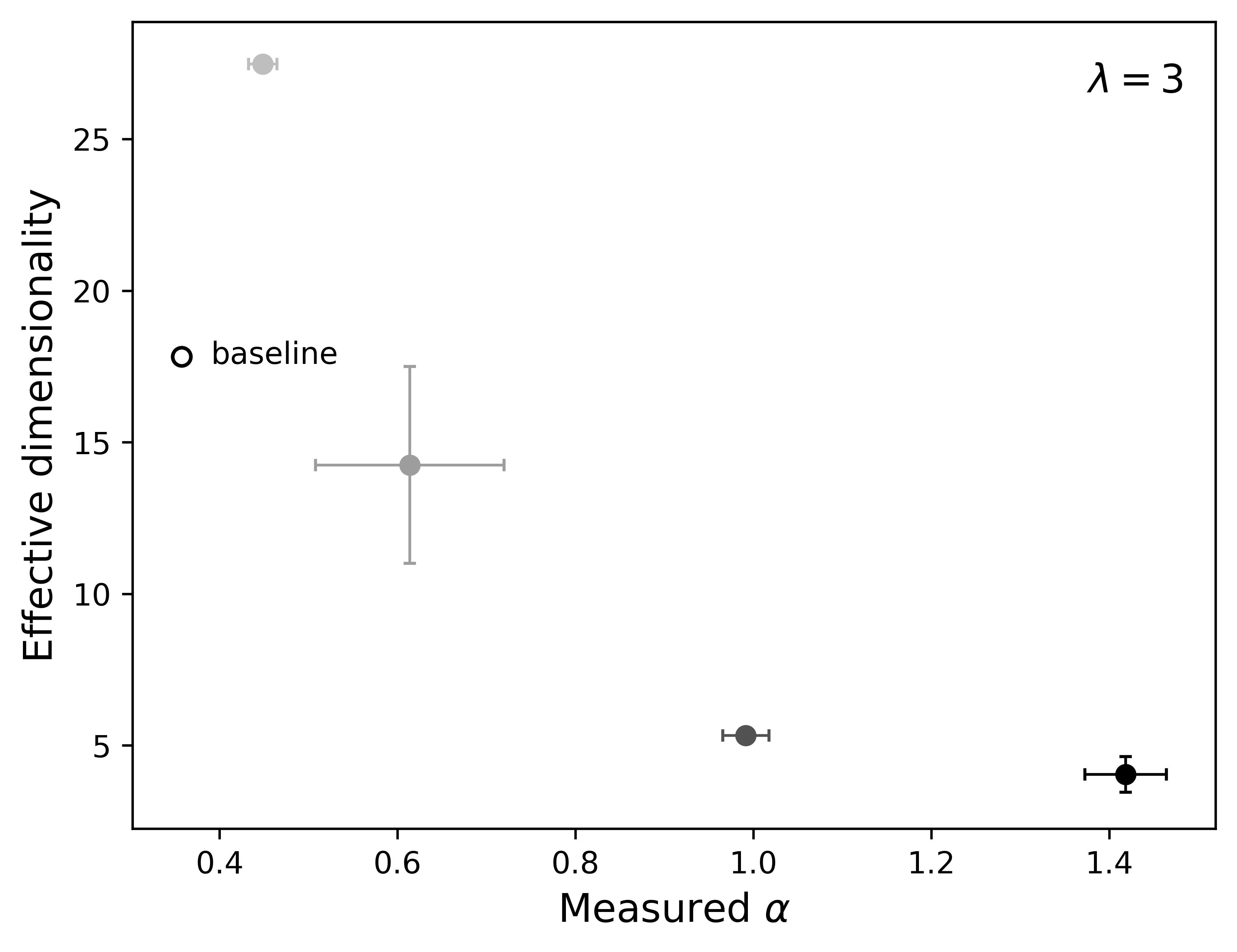}
    \caption{Effective dimensionality as a function of the measured spectral exponent $\alpha$. The white point represents the unregularized baseline. Error bars represent the median absolute deviation.}
    \label{fig:effective_dimensionality}
\end{figure}

For intermediate measured $\alpha$, the forward–reverse asymmetry was nearly eliminated within the common-unit subspace, with forward and reverse predictivity reaching comparable levels. This result suggests that steering representational geometry preferentially reorganizes model representations already shared with—and predictive of—neural activity, improving their linear recoverability. Consistent with this, effective dimensionality decreased as the measured $\alpha$ increased (Fig.~\ref{fig:effective_dimensionality}), indicating that steering representational geometry progressively concentrates representational variance into fewer effective dimensions. Together, these findings suggest that improved bidirectional representational alignment is accompanied by a reorganization of the shared representational subspace and a broader concentration of population-level variance. Although spectral regularization likely induces multiple changes to the learned representations, these observations are consistent with our hypothesis that representational geometry contributes to improved bidirectional predictivity.

\section{Discussion}
In this study, we developed a computational framework for steering representational geometry and showed that doing so influences bidirectional representational alignment between biological and artificial neural networks. Steering the spectral geometry of the learned representations increased reverse predictivity with only modest reductions in forward predictivity, reducing asymmetry between the two directions and improving overall bidirectional predictivity. These changes were associated with a reorganization of the shared representational subspace and reduced effective dimensionality, providing insight into potential mechanisms underlying the improved representational alignment.

These findings extend previous work on representational alignment. For example, task-optimized models have been shown to develop representations that predict neural activity despite not being explicitly optimized using neural data \cite{og_predictivity,Schrimpf2020IntegrativeBenchmarking,ssl}. Recently, \cite{reverse} showed that this alignment is asymmetric, with model representations predicting neural responses much better than neural responses predict model representations. Our research demonstrates that this asymmetry is not fixed, but can be reduced when representational geometry is steered during learning. Increasing the measured spectral exponent primarily improved reverse predictivity, while forward predictivity was more strongly associated with the regularization strength. Within the common representational subspace, forward and reverse predictivity became approximately symmetric at intermediate spectral exponents. Together, these findings demonstrate that forward--reverse asymmetry can be systematically steered by interventions that change the organization of the learned representations and are consistent with representational geometry contributing to this asymmetry.

Our findings also extend previous work on representational geometry. For example, research has shown that neural representations exhibit characteristic spectral geometry \cite{geom}, while the spectral properties of artificial neural networks have been associated with representation quality and external behavior \cite{lossfunc}. Building on previous research \cite{lossfunc}, our framework applies controlled spectral steering to systematically study its effects on bidirectional representational alignment.

Notably, we found that bidirectional alignment continued to improve beyond the previously reported spectral exponent of approximately $\alpha=1$ in the visual cortex \cite{geom}, indicating that matching this spectral exponent is not itself sufficient to maximize alignment. Increasing spectral exponents were also associated with reduced effective dimensionality and improved reverse predictivity within the common representational subspace, consistent with previous research linking effective dimensionality to bidirectional representational alignment \cite{reverse}. More broadly, our research establishes representational geometry as a steerable property of learned representations, providing a framework for studying its role in bidirectional representational alignment.

One limitation of our work is that the spectral exponent provides only one characterization of representational geometry. Representations with similar spectral exponents may differ in other geometric or statistical properties, and spectral regularization may alter properties of the learned representations beyond the spectral exponent itself. We found that forward predictivity varied with regularization strength even among models with similar measured spectral exponents, indicating that the observed changes in bidirectional alignment cannot be attributed exclusively to the spectral exponent. Accordingly, the measured spectral exponent should be interpreted as one property of learned representations associated with changes in alignment rather than as the sole causal determinant of such changes. Another limitation is the scope of our evaluation. As a proof-of-concept, we demonstrated our framework using a self-supervised contrastive ResNet-50 model and evaluated alignment at one model layer using a single neural benchmark. Whether the observed trends generalize across other architectures, learning objectives, model layers, and datasets remains to be determined.

These limitations motivate future research. First, additional properties of representational geometry beyond the spectral exponent should be studied to determine which properties most strongly influence bidirectional representational alignment. Developing more selective approaches for steering representational geometry, including adaptive or scheduled spectral regularization, could also help disentangle the effects of representational geometry from those introduced by regularization strength. The generality of our findings should also be evaluated across other architectures, learning objectives, model layers, representations, and datasets \cite{wong2026,zbaranska2026}. Finally, our computational framework could be extended to study the representational alignment between pairs of deep learning models \cite{modelstitching}. This would allow for more controlled investigations of how representational geometry influences alignment and could inform broader approaches for understanding and interpreting learned representations \cite{internaldistillation,justin2026}. 

\section{Acknowledgments}
We wish to thank members of the Machine Intelligence Lab for their insightful discussions and feedback. We dedicate this research to the students and researchers in Ukraine. Their resilience and unwavering commitment to learning and education continue to serve as a beacon of hope and inspiration to the global academic community.

\vfill

\end{document}